\documentclass{llncs}
\usepackage{times}
\usepackage{graphicx}
\usepackage{latexsym}
\usepackage{amsmath}
\usepackage{tabu}

\begin{document}
\bibliographystyle{splncs}
\title{Psychologically based Virtual-Suspect for Interrogative Interview Training}

\author{
    Moshe Bitan \institute{Bar Ilan University, Israel}  
    \and
    Galit Nahari  \institute{Bar Ilan University, Israel}
    \and
    Zvi Nisin  \institute{Israeli Police, Israel}
    \and
    Ariel Roth  \institute{Bar Ilan University, Israel}
    \and
    Sarit Kraus  \institute{Bar Ilan University, Israel} 
}

\maketitle

\begin{abstract}
  In this paper, we present a \textit{Virtual-Suspect} system which can be used to train inexperienced law enforcement personnel in interrogation strategies. The system supports different scenario configurations based on historical data. The responses presented by the Virtual-Suspect are selected based on the psychological state of the suspect, which can be configured as well. Furthermore, each interrogator's statement affects the Virtual-Suspect's current psychological state, which may lead the interrogation in different directions. In addition, the model takes into account the context in which the statements are made. Experiments with 24 subjects demonstrate that the Virtual-Suspect's behavior is similar to that of a human who plays the role of the suspect.

\end{abstract}

\section{Introduction}
One of the most important tasks of a police officer is to interrogate suspects.
The main role of the police interrogation is to encourage the suspect to inadvertently incriminate himself and to collect further information regarding the case. 
Often, the physical evidence available in a given investigation is scarce, which makes the interrogation stage much more critical.
Therefore, police cadets undergo rigorous interrogation training.

One of the leading training techniques is one-on-one interrogation simulation sessions. In these personal sessions, a trained instructor conducts an interrogation simulation with the trainee acting as the investigator. In order to prepare the inexperienced law enforcement personnel for real-world investigations, the instructor must develop a scenario based on real cases. Furthermore, the instructor or a hired actor plays the role of the suspect and portrays different personalities based on the corresponding scenario. 
The interrogation uses 
the \textit{Investigative Interviewing} method and is based on an interrogation plan starting with a relaxed atmosphere of acquaintance, through obtaining the suspect's version of the event, verifying the suspect's alibi and finally introducing tough questions and in some cases making accusations against the suspect. Interestingly, interrogators often cycle back to one or more of these steps according to the suspect's state of mind. This training technique has long served its goals and has proven efficient and effective. Unfortunately, this time consuming technique requires experienced instructors or actors. Furthermore, the training sessions are carried out on a one-trainee-at-a-time basis, and are therefore expensive.

Virtual-Suspect systems may offer a solution for this. 
The advantages of a computer interrogation training system are fairly straightforward.
First, the system can provide a large number of scenarios and satisfy greater control over the Virtual-Suspect's personality. Second, multiple cadets can train simultaneously at their convenience with an instructor monitoring their progress. Lastly, detailed reports can be provided as well as transcripts of an interrogation which can facilitate instructors in monitoring the training.

In this paper, we present a Virtual-Suspect system that was developed in collaboration with criminology researchers, experienced criminal psychologists and the Israeli Police Department (IDP). The system supports different scenario configurations based on real cases. The responses presented by the Virtual-Suspect are selected based on the psychological state of the suspect, which can be configured as well. Furthermore, each interrogator's statement affects the Virtual-Suspect's current psychological state, which may lead the interrogation in different directions. In addition, the model takes into account the context in which the statements are made.

In addition, an experiment was conducted comparing the system's responding mechanism with that of a human instructor. The experiment was divided into two phases. In the first phase, interrogation simulations were conducted either using a human instructor, our system or a baseline random system. In the second phase, participants were asked to read the transcripts of these simulations and answer a series of questions. The results suggest that humans have difficulty differentiating between simulations generated by our system and those of a human instructor. To conclude, the cost efficient \textit{Virtual-Suspect} system can be used around the clock to train inexperienced law-enforcement personnel in a variety of realistic interrogations.

\section{Related Work}

The effectiveness of Virtual-Humans in training systems has been studied extensively in computer science \cite{hubal2001interactive,kenny2007building,parsons2008objective}.
Anderson et al. \cite{anderson2013tardis} presented the \textit{TARDIS}, an interview coaching system for training  scenario-based interactions. The framework incorporates a \textit{Non-Verbal Behavior Analyzer (NovA)} sub-module which can recognize several lower level social cues, e.g. hands-to-face, postures, leaning forward or backward, among others. Luciew et al. \cite{luciew2011finding} presented an immersive interrogation learning simulation specifically designed for training law enforcement personnel in interviewing children who were victims of sexual abuse and interrogating suspects on that matter. The study puts a strong emphasis on the interpretation of nonverbal behavior by police officers during an interview or interrogation. 

Other researchers have focused their attention on the Virtual-Suspect's psychological model.
Roque and Traum \cite{roque2007model} categorized three compliancy levels: compliant, reticent and adversarial. A compliant suspect provides semi-useful information when asked. A reticent suspect provides neutral information and evades any questions about sensitive topics. An adversarial suspect provides deceptive or untruthful responses. Similar to this compliancy categorization, the Virtual-Suspect's internal-state as presented in this paper integrates a continuous compliancy parameter which, in combination with other parameters, determines the suspect's response. A formal description of these parameters is presented later in section \ref{ssec:Personality}. Olsen at el. \cite{olsen1997interview} presented a police interrogation simulation system teaching police cadets to build a rapport with the suspect while maintaining professionalism, to listen to verbal cues and to detect important changes in both verbal and nonverbal behavior. The Virtual-Suspect, a male loan officer, is accused of stealing from an ATM. The presented response model is based on the suspect's internal-state, comprising of the suspect's mood and the rapport that has been established between the suspect and interrogator. The system presented in this paper extends Olsen's internal state model and introduces a three-dimensional internal state. Furthermore, the system supports multiple configurable interrogation scenarios typically based on real cases (see section \ref{ssec:IntConfig}). Lastly, Bruijnes et al. \cite{bruijnes2014virtual} presented a Virtual-Suspect response model that can portray a variety of unique personalities. The system is based on static personality parameters as well as a dynamic interaction state. Similar to the experiment presented in this paper, Bruijnes conducted an experiment measuring a human being's ability to distinguish between different Virtual-Suspect personalities.

\section{Methodology}\label{sec:Method}

The Virtual-Suspect simulation system proposed in this paper enables the simultaneous simulation of multiple interrogation cases and supports different Virtual-Suspect personality configurations. Furthermore, a simulated interrogation can be unfolded in many different directions depending on the trainee's behavior. In order to facilitate this flexibility, the system is comprised of two major components; The first component is the \textit{Simulation Configuration} and the second component is the \textit{Virtual-Suspect Response Model}.
In the following sections a detailed description and a formal representation of these two components and their sub-components are presented.

\subsection{Simulation Configuration} \label{ssec:SimConfig}

The \textit{simulation configuration} provides a high level of control over the simulated interrogation. First, the interrogation scenario configuration enables the development of complex investigation simulations, typically based on real cases. Second, the system supports a customizable template database of statements and responses. Third, the Virtual-Suspect's personality, which is based on Eysenck's PEN Model \cite{eysenck1990biological}, can be configured. These three levels of control enable the interrogation trainer to simulate a wide range of interrogation scenarios.

\subsubsection{Interrogation Scenario Configuration} \label{ssec:IntConfig}

The scenario configuration is implemented via a personal information database and an event database. The personal information database stores the Virtual-Suspect's \textit{age, marital status, spouse, children, last known address, occupation, income, place of employment, known acquaintances} and any other relevant personal data. 
The event database contains information about occurrences that may or may not have happened to the Virtual-Suspect. These occurrences (or events) contain multiple types of information which relate to the occurrence, that is the \textit{location, time, date, activity, participants, objects} and \textit{means of transportation}. For example, on the 24th of December 2013 at 8:30 pm, the Virtual-Suspect and his wife dined at their residence.

An event can be either truthful or false. A truthful event is an event that actually happened to the Virtual-Suspect. On the other hand, a false event is an occurrence that the Virtual-Suspect may use in order to give a deceptive response, and it is usually associated with providing \textit{an Alibi} or \textit{Legal Access}. Therefore, events are labeled with one of the following mutually exclusive labels: \textit{Criminal, Alibi, Legal Access or Neutral}. A \textit{Criminal} event is the offense details allegedly committed by the Virtual-Suspect. \textit{An alibi} event is a defense used by the Virtual-Suspect attempting to prove that he or she was in some other place at the time that the alleged offense was committed. \textit{A Legal Access} event is another defense used by the Virtual-Suspect attempting to provide a legal explanation for its presence in the alleged offense location, association with an alleged accomplice to the offense or access to alleged stolen items. Lastly, a \textit{Neutral} event is any event other than the above mentioned events. This event database models the long-term memory of the Virtual-Suspect. However, in real-world cases, the police investigators are provided with only a partial view of what actually happened. To simulate this, at the onset of a simulation, the trainee is provided with a case file which consists of the information known to the police as well as the collected forensic evidence.

In real-world investigations, some topics can trigger a more emotional response than others. In order to simulate this behavior, a dedicated label was introduced to the system. The $hot$ label is an indication that a personal detail, an event or any detail pertaining to an event has a profound effect on the Virtual-Suspect when asked about the matter. It is important to note that an event or an event detail can be labeled as $hot$ regardless of the event label. For example, if the Virtual-Suspect's spouse's information is labeled as $hot$, an inquiry about a \textit{Neutral} event where the Virtual-Suspect and his wife dined at their residence might have a profound effect on the Virtual-Suspect's state of mind. A more in-depth explanation of the $hot$ label and its effects is provided later in section \ref{ssec:ResponseModel}.

\begin{figure}[h]
\begin{center}
\includegraphics[width=\textwidth]{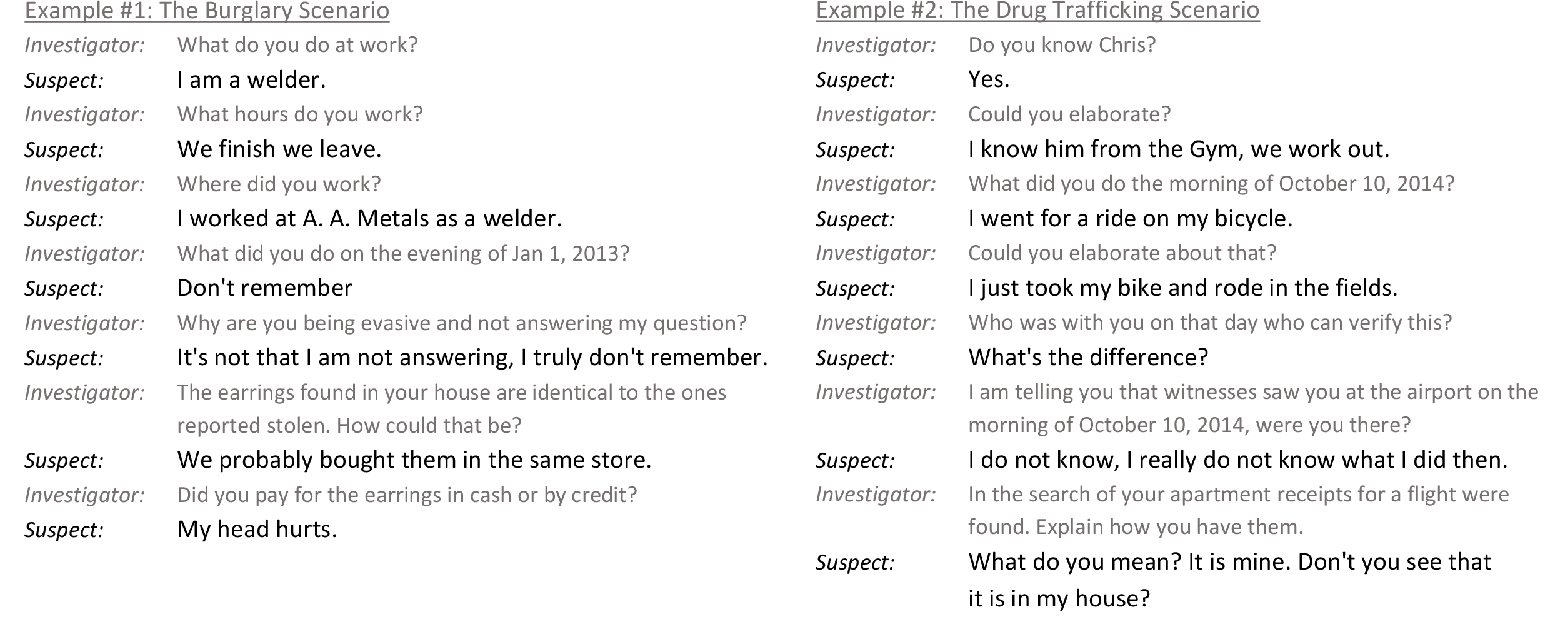}
\end{center}
\caption{Simulation Transcripts}
\label{fig:sims}
\end{figure}

To demonstrate the flexibility of the Interrogation Scenario Configuration, two scenario configuration examples are presented. The first example is the scenario used in the experiment (see section \ref{sec:ExpMethod}). The second example was configured for the purpose of demonstrating the ease of use of the system. Furthermore, the complex scenario was configured in less than two hours.
Simulation transcripts, conducted with these two scenario configurations, can be viewed in Figure \ref{fig:sims}.

\begin{example}\label{ex1}

The interrogation scenario chosen for the experiment is based on an actual burglary case from early 2013. A 46-year-old married welder was charged with breaking and entering into a private residence and stealing a pair of valuable earrings and a laptop computer. Forensic investigation led to the finding of the suspect's fingerprints on a window ledge. As a result, a search warrant was granted, leading to the discovery of the earrings hidden in the suspect's residence. Finally, the suspect was arrested and escorted to the interrogation room.

\end{example}

\begin{example}

The second interrogation scenario is also based on a real case from late 2014.
Bob, a 28-year-old married delivery man was charged with drug trafficking. 
The suspect's contact information was found on a sim-card belonging to Chris, a second offender. Chris was arrested at the airport as he was trying to leave the country. An on-site search led to the discovery of 2Kg of an illegal substance in his possession. Furthermore, witnesses have testified to seeing Bob's car at the airport when he allegedly purchased the plane tickets for Chris. A further forensic investigation led to the finding of the suspect's fingerprints on a ticket purchasing booth. As a result, a search warrant was granted, leading to the discovery of the ticket receipts hidden in the suspect's residence. Finally, the suspect was arrested and escorted to the interrogation room.

\end{example}

\subsubsection{Interrogative Interviewing}\label{ssec:QnA}

The interrogation simulation is based on a standard Interrogative Interviewing model. Unlike other systems that support a single static scenario \cite{olsen1997interview,bruijnes2014virtual},
the Virtual-Suspect system supports multiple configurable scenarios. Therefore, the Virtual-Suspect system provides the ability to configure statement and response templates. Each statement or response template is comprised of a static text portion and a dynamic input fields portion. The input fields portion is filled out by the investigation trainee during the simulation. For example, ``Where were you on 01/01/2013?'' is an instance of the ``Where were you on \textit{[Date]}?'' statement template. During the simulation, the trainee selected the statement template from the templates database and entered the date `01/01/2013'.

The association of a statement template to a response template is a \textit{Many-to-Many} relation. In other words, a statement can have multiple response templates associated with it. Symmetrically, a response template can be associated with multiple statement templates. For instance, the statement ``How are you?'' can be associated with ``I am feeling well'' as well as with ``I am a little agitated''. Likewise, the response template ``I don't remember'' can be associated with the statement template ``Where were you on \textit{[Date]}?'' as well as to ``Where did you purchase these \textit{[Objects]}?''.

During a simulation, the trainee selects a statement from the statement-response templates database, fills in the input fields and sends the statement to the Virtual-Suspect. Consequently, the Virtual-Suspect's response model extracts all associated responses from the templates database, fills in the required input fields from the databases, and selects an appropriate response from the associated responses set.

As described above, the $hot$ label indicates that an information detail (i.e. a field) has a profound effect on the Virtual-Suspect when asked about the matter. To simulate this behavior, the $hot$ label is propagated at runtime to the Interrogative Interviewing model. A response or a statement is marked as $hot$ if one or more of its input field values is labeled $hot$ in the Scenario Database. In addition, a statement is also marked $hot$ if any of its associated responses is marked as $hot$. Lastly, if a statement is not marked as $hot$ it is considered to be $cold$. Formally,
the indicator function $\delta^{hot}$ indicates that a given statement $q$ is marked as $hot$:
\begin{equation}
	\delta^{hot}(q)=
\begin{cases}
	1 ,& \text{if } \exists field \in \cup_{r \in R_q} F^r  \cup F^q \text{ where } field \text{ is labeled as } hot\\
	0,          	& \text{otherwise}
\end{cases} 
\end{equation} Where $q$ denotes the statement, $r$ denotes a response, and $R_q$ is the associated responses set, $F^q$ and $F^r$ are the set of input fields for the templates $q$ and $r$, respectively.

A statement directly affects the internal-state of the Virtual-Suspect.
However, not all statement's subjects have an equal effect on the psychological state of the suspect. Therefore, the weight of the effect on the internal-state of each statement template can be tuned individually during the configuration. For each statement template $q$ two weight vectors are defined. The first, denoted by $w^{hot}_q$, determines the effect when the statement is marked as $hot$. The second, denoted by $w^{cold}_q$, determines the effect when the statement is marked as $cold$ (i.e. not $hot$). An effect weight component can be either 1, -1, or 0 representing a positive, a negative and a neutral effect, respectively. Since the internal-state is represented using a three-dimensional vector (see section \ref{ssec:ResponseModel}), these weight vectors are three-dimensional as well. Formally, the vectors can be described as:  $w^{hot}_q, w^{cold}_q \in \{-1, 0, 1\}^3$. These vectors were determined separately by two experts. In case of a disagreement, a discussion was made to resolve it.

\subsubsection{Personality Profile}\label{ssec:Personality}

The Virtual-Suspect personality profile consists of two intertwined components. The first is the Virtual-Suspect's internal-state which is based on Eysenck's \textit{PEN} Model of personality \cite{eysenck1990biological}. The second is the volatility of the Virtual-Suspect's personality, that is, the degree to which a statement affects the internal-state.

The Virtual-Suspect's internal-state is based on Eysenck's \textit{PEN} Model of personality \cite{eysenck1990biological}. More specifically, Eysenck's \textit{PEN} Model consists of the \textit{Psychoticism, Extraversion and Neuroticism} personality traits where \textit{Psychoticism} refers to a personality pattern typified by aggressiveness and interpersonal hostility. \textit{Extraversion} tends to be manifested by outgoing, talkative, energetic behavior, whereas introversion is manifested by a more reserved and solitary behavior. \textit{Neuroticism} is a fundamental personality trait in the study of psychology characterized by anxiety, fear, moodiness, worry, envy, frustration, jealousy, and loneliness. Individuals who score high for neuroticism are more likely than the average human to experience such feelings as anxiety, anger, envy, guilt, and depression. Formally, the internal-state is a three-dimensional vector denoted by $s = (s_1, s_2, s_3) \in \Re^3$ where the $s_1, s_2$ and $s_3$ components correspond to the \textit{Psychoticism, Extraversion and Neuroticism} personality traits, respectively. The initial value of the internal-state vector, denoted by $s_0$, is configured by the interrogation simulation supervisor (i.e. trainer) and is an integral part of configuring an interrogation simulation.

The internal-state changes during an interrogation simulation in order to reflect the effect that the interrogator's statements have on the Virtual-Suspect. The degree to which the internal-state varies is determined by the second component, that is, the personality \textit{Volatility} parameter. More specifically, the \textit{Volatility} parameter, denoted by $\sigma$, is a non-negative, three-dimensional weight vector $\sigma \in \Re_+^3$ that determines the pace at which the internal-state changes for any given statement. For instance, configuring the \textit{Volatility} parameter, $\sigma$, to a value of $(1, 0, 0)$ will cause the interrogator's statements to only effect the \textit{Psychoticism} component of the internal-state vector. On the other hand, setting the $\sigma$ parameter to a high value such as $(3, 3, 3)$ will cause the internal-state vector to change dramatically after every statement. Consequently, the Virtual-Suspect will react in an erratic and unstable manner. Although this   \textit{Volatility} parameter is configured in the Simulation Configuration part, its main function is in updating the internal-state vector which is part of the Response Model. A more detailed description of the \textit{Volatility} parameter function will be presented later in section \ref{ssec:StateUpdate}.



\subsection{Response Model}\label{ssec:ResponseModel}

The Virtual-Suspect \textit{Response Model} generates the Virtual-Suspect's response for a given inquiry, i.e. an investigator's statement. The model is comprised of four cognitive components. The first two components are the long- and short-term memory components. The long-term memory gives the Virtual-Suspect access to the Interrogation Scenario database to recall personal information and past events. Equally important, the short-term memory gives the Virtual-Suspect the ability to respond to follow-up statements. The third component, and arguably the most important, is the Virtual-Suspect's internal-state update mechanism which determines the way an investigator's statement affects the Virtual-Suspect's internal-state. Finally, all of the above components are combined to produce the response of the Virtual-Suspect (see Section \ref{ssec:responseSelection}).

\begin{figure}[h]
\begin{center}
\includegraphics[width=\textwidth]{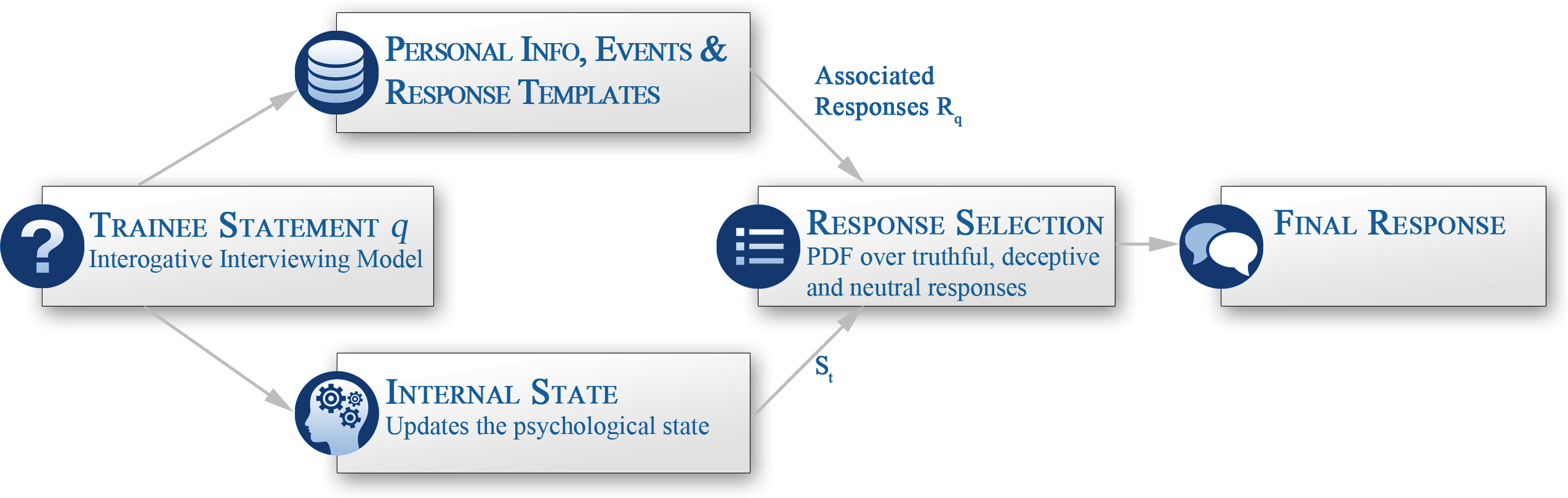}
\end{center}
\caption{Response Model}
\label{fig:ResponseModel}
\end{figure}

As can be viewed in Figure \ref{fig:ResponseModel}, once a statement $q$ is selected by the trainee, the response model generates a response using the following steps. First, the model extracts the associated responses set$R_q$ from the \textit{Interrogative Interviewing Database}. Second, the long- and short-term memory are used to populate all input fields for each response in the associated responses set $R_q$. Next, the \textit{Response Model} updates the internal-state vector. Finally, the updated internal-state vector is used to select the most appropriate response from the associated responses set $R_q$. 

\subsubsection{Long- and Short-Term Memory}\label{ssec:Memory}

The long-term and short-term memory components' main purpose is to populate the input fields for the associated responses set $R_q$. Given a statement $q$, the \textit{Response Model} determines if the statement is a follow-up statement, refers to a new event or is a generic statement (e.g. how are you feeling today). If the statement refers to an event, the model extracts all relevant information from its long-term memory (i.e. the Scenario Database) and stores it in its short-term memory. Lastly, the \textit{Response Model} populates all input fields for each response in the associated responses set $R_q$ using the event information stored in the short-term memory and the personal information database.

\subsubsection{Internal State Vector Update Mechanism}\label{ssec:StateUpdate}

The \textit{Internal-State Vector Update Mechanism} is executed for every statement. However, not all statement's subjects have an equal effect on the psychological state of the suspect, as mentioned in section \ref{ssec:QnA}.
During a simulation, for every statement $q$ at time $t$ the internal-state vector is calculated using the following equation:
\begin{equation}\label{eq:Update}
s_{t} = s_{(t-1)} + \sigma \cdot \big( \delta ^{hot}(q)\cdot w^{hot}_{q} + \big(1 - \delta ^{hot}(q)\big)\cdot w^{cold}_{q}\big)
\end{equation}
Where $\sigma$ is the \textit{Volatility} parameter, $\delta ^{hot}(q)$ indicates if a statement $q$ is marked as $hot$ and $w^{hot}_{q}$, $w^{cold}_{q}$ are the statement effect weight vectors. It is important to note that the \textit{Volatility} parameter $\sigma$ is configured in the Personality Profile configuration phase and remains constant for the duration of the simulation (see section \ref{ssec:Personality}).

\subsubsection{Response Selection}\label{ssec:responseSelection}

\begin{figure}[h]
\begin{center}
\includegraphics[width=\textwidth]{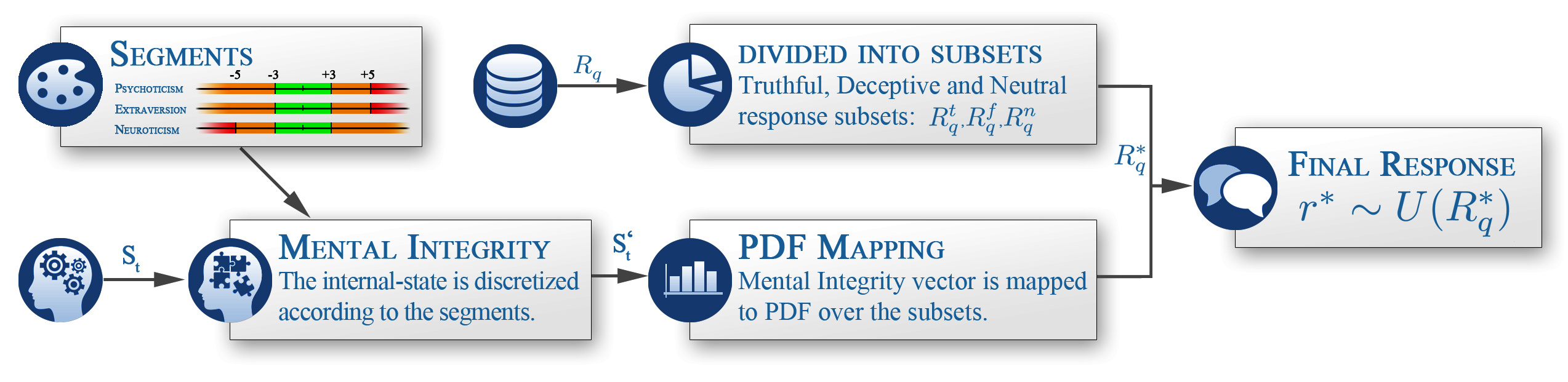}
\end{center}
\caption{Response Selection}
\label{fig:responseSelection}
\end{figure}

The final phase in the response model is the response selection phase. In this phase the internal-state vector is used to select the output response, denoted by $r^*$, from the associated responses set $R_q$. More specifically, the internal-state determines what type of response to select; a truthful, deceptive or neutral response. As can be seen in Figure \ref{fig:responseSelection}, the response selection process consists of four steps. 

First, the associated responses set, $R_q$, is divided into three subsets: : separate subsets with truthful, false and neutral responses. The truthful responses subset, denoted by $R^t_q$, contains all responses that relate to truthful events, that is events that actually happened to the Virtual-Suspect. The false responses subset, denoted by $R^f_q$, contains all deceptive responses, that is responses related to \textit{Alibi} or \textit{Legal-Access} events. The neutral responses subset, denoted by $R^n_q$, contains all general responses, for instance, ``I do not recall'', ``I cannot remember'' and ``Let me be''.

In the second step, the internal-state vector components are color-coded and then discretized to the \textit{Mental Integrity} vector, denoted by $s'$. First, the value range of each internal-state vector component is divided into three sections, denoted by \textit{$I_{g}$, $I_{o}$} and \text{$I_{r}$}, where each segment is color-coded to \textit{green, orange} and \textit{red}, respectively. The color-coding expresses the level of the Virtual-Suspect's \textit{Mental Integrity}. The \textit{green} color-coded section, $I_g$, indicates that the Virtual-Suspect is mentally stable. The \textit{orange} color-coded section, $I_o$, indicates that the Virtual-Suspect is moderately stable. Lastly, the \textit{red} color-coded section, $I_r$, indicates that the Virtual-Suspect's mental integrity is compromised. Formally, let $s$ be the internal-state vector, and the \textit{Mental Integrity} vector $s'\in{\{0,..., 3\}}^3$ components can be calculated as follows:
\begin{equation}
	{s'}_1=\sum_{i=1}^{3}\delta\big(s^i\in I_{g}\big)\;, \; \; \;
	{s'}_2=\sum_{i=1}^{3}\delta\big(s^i\in I_{o}\big)\;, \; \; \;
	{s'}_3=\sum_{i=1}^{3}\delta\big(s^i\in I_{r}\big)
\end{equation}
where $\delta$ is the standard indicator function (i.e. equals 1 if the condition is true, otherwise equals 0). 
The sections were carefully calibrated by police psychologists. For \textit{Psychoticism} and \textit{Extraversion}, the sections were set to $I_{g} = [-3, 3]$,  $I_{o} = (-\infty, 3)\cup(3, 5]$ and  $I_{r} = (5, \infty)$.
For \textit{Neuroticism}, the sections were set to $I_{g} = [-3, 3]$,  $I_{o} = [-5, -3)\cup(3, \infty]$ and  $I_{r} = (-\infty, -5)$. Figure \ref{fig:responseSelection} provides a visual representation of the segments.

The main objective of the Virtual-Suspect is to lead the interrogator to believe he or she has nothing to do with the alleged offense. Therefore, if the Virtual-Suspect is guilty it will try to deceive the interrogator when asked about topics related to the \textit{Criminal} event. However, if the Virtual-Suspect's mental integrity is compromised, there is a higher probability that the Virtual-Suspect will inadvertently choose a response that does not serve its goals. Furthermore, when asked about a $hot$ topic that is not related to the \textit{Criminal} event, the Virtual-Suspect will try to avoid the question and there is a high probability that it will respond with an elusive response. In particular, this context-dependent behavior is modeled in the third step. More specifically, in the third step, the \textit{Mental Integrity} vector and the contextual event label are mapped to a probability distribution function over the three response subsets, i.e. $R^t_q$, $R^f_q$ and $R^n_q$, determining what type of response will be selected. For example, if the Virtual-Suspect is presented with a statement regarding the \textit{Criminal} event and the Virtual-Suspect is mentally stable, i.e. $s'=(3,0,0)$, then there is a high probability of responding with a deceptive response. Specifically, the corresponding distribution probability function will be set to $p(r^*\in R^f_q) = 1$, $p(r^*\in R^t_q) = 0$ and $p(r^* \in R^n_q) = 0$. On the other hand, for the exact same statement, if the mental integrity is compromised, i.e. $s'=(0,0,3)$, indicating that the Virtual-Suspect's cognitive capacity is diminished, then there is a low probability of responding with a deceptive response. Specifically, the corresponding distribution probability function will be set to $p(r^*\in R^f_q) = 0.1$, $p(r^*\in R^t_q) = 0.5$ and $p(r^* \in R^n_q) = 0.4$. Lastly, the associated responses subset is randomly selected using the corresponding distribution probability function and is denoted by $R^*_q$. 

Finally, in the fourth step, the output response, $r^*$, is randomly selected from the previously selected subset, $R^*_q$, using a uniform distribution function, i.e. $r^* \sim U(R^*_q)$. For instance, if the deceptive responses subset was selected, i.e. $R^*_q = R^f_q$, and $R^f_q$ contained five deceptive responses, then all five deceptive responses would have an equal chance to be selected as the final response of the Virtual-Suspect.

\section{Experiment}\label{sec:ExpMethod}

\subsection{Methodology}\label{sec:ExpMethodology}
A meticulous experiment was conducted in order to measure the effectiveness of the Virtual-Suspect response model (RMVS) as compared to a human. Participants were asked to read transcripts of three interrogation simulations and answer a series of questions. As described in section \ref{ssec:IntConfig}, the interrogation scenario chosen for the experiment is based on an actual burglary case from early 2013 (see Example \ref{ex1}). In addition, the Virtual-Suspect's personality profile was chosen to represent a moderately calm individual. More specifically, the initial internal-state vector was set to $s_0=(0, 0, -3)$ for the \textit{Psychoticism, Extraversion and Neuroticism} personality traits, respectively. In addition, the \textit{Volatility} parameter was set to $\sigma = (0.5, 0.5, 0.5)$. In the following sub-sections, a detailed description of the experiment's methodology is presented. 


Prior to the experiment, three different Virtual-Suspect models were simulated. In the first, a human instructor acted as the suspect. The second was the Virtual-Suspect response model, i.e. RMVS. In the third, a randomized response selection mechanism was used as a baseline. 
%
%
%
In the experiment, 24 participants, ranging in age from 20-30, were asked to carefully read the transcripts of the three simulations and answer a series of questions. 12 of the participants were males and 12 were females. The order in which the simulations' transcripts were presented to participants was selected with great care in order to ensure that the results remain order independent. Consequently, each possible order of the three transcripts was presented to four participants, two males and two females.

Our hypothesis was that humans will find it difficult to differentiate between a human trainer and the Virtual-Suspect response model. However, the randomized response selection baseline mechanism will be distinguishable. The next section provides an in-depth presentation of the results following a discussion of their importance. 

\subsection{Results}

\begin{figure}[h]
\begin{center}
\includegraphics[width=\textwidth]{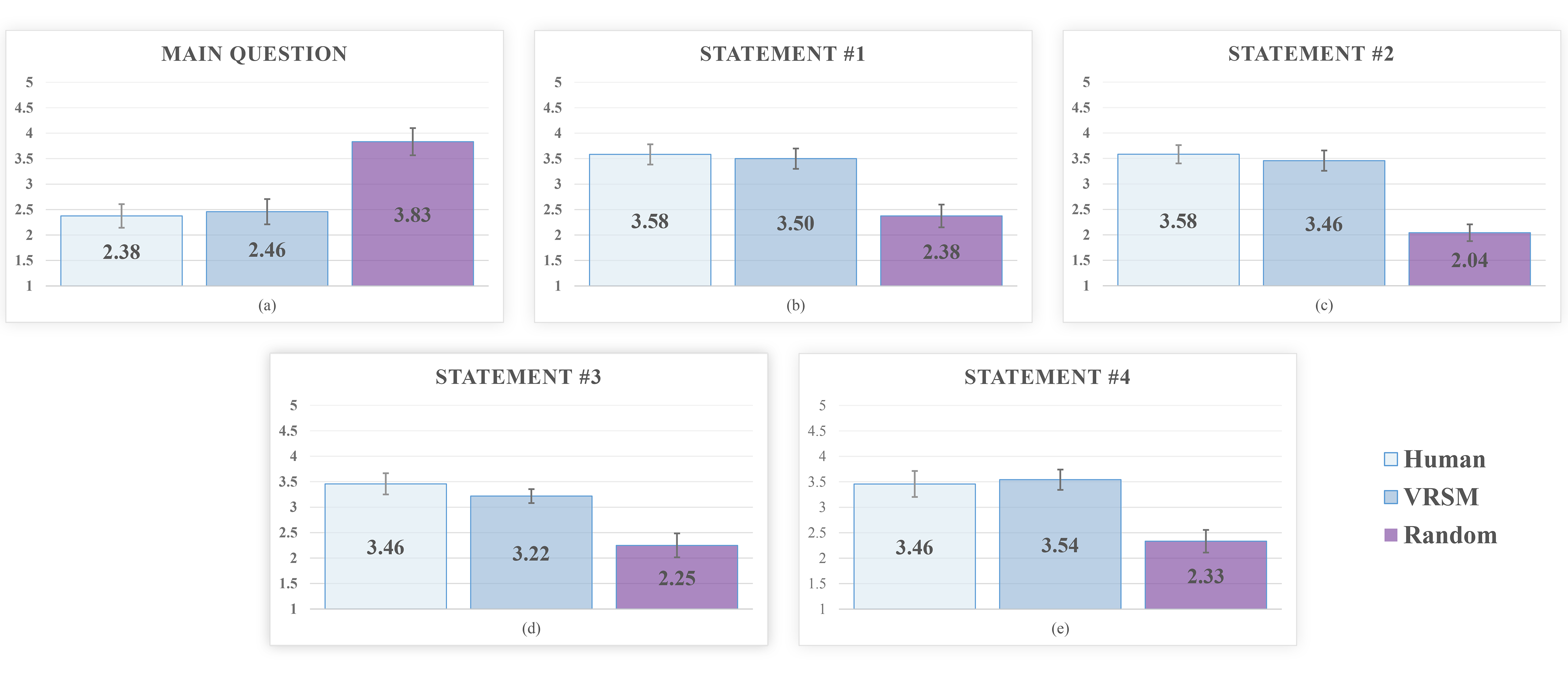}
\end{center}
\caption{Experiment's Results}
\label{fig:results}
\end{figure}

As described in section \ref{sec:ExpMethodology}, participants were asked to carefully read the three simulation transcripts and answer a series of questions. Since the experiment's hypothesis was that humans will find it difficult to differentiate between a human and the \textit{RMVS}, the participants were presented with the following question: \\
\textit{``The suspect's responses in the transcripts have been selected from an existing response database. In some of the transcriptions, the responses have been selected by a human participant, and in the others by a computer. Would you say that in this current transcription the responses were chosen by the human participant or by the computer?''.}\\
The participants were asked to specify their belief in a \textit{Likert}-type scale ranging from one to five, one being that the suspect's responses have been selected by a human participant and five being that the suspect's responses have been selected by a computer. The results are presented in Figure \ref{fig:results} (a). The average score for the Human condition is $2.38$, for RMVS is $2.46$ and for the Random baseline it is $3.83$ (lower is better, i.e. perceived as more human). To analyze the results, repeated measures ANOVAs were conducted to compare the three conditions, i.e. Human, Random and RMVS. The resulting ANOVAs are $F(2, 22) = 8.15$, partial $\eta2= 0.57$, $p  < 0.002$. As hypothesized, the RMVS were rated significantly better, i.e. more human like, than the Random baseline (using paired t-test $p<0.01$). Similarly, the Human transcript was rated significantly better than the Random baseline as well (using paired t-test $p<0.01$). Furthermore, the Human transcript was rated slightly better than the RMVS. However, this small difference is not statistically significant. 

To further support our hypothesis, the results of the following four statements are presented:
\textit{
\begin{enumerate}
    \item The simulated suspect behaves as expected of a real suspect.
    \item The suspect's responses are relevant to the interrogator's statements.
    \item The interrogation is coherent.
    \item The interrogation is realistic.
\end{enumerate}
}
The participants were asked specify their level of agreement or disagreement on a standard five level \textit{Likert}-type scale for these four statements. The \textit{Likert}-type scale consists of the following five levels: 1. Strongly disagree, 2. Disagree, 3. Neither agree nor disagree, 4. Agree and 5. Strongly agree. The averages and standard deviation results are presented in Figure \ref{fig:results} (b) to (e). Running repeated measures ANOVAs on the three conditions yielded similar results to the results for the main question. In addition, as can be seen in the figure, the RMVS performed significantly better than the Random baseline (using a paired t-test $p<0.01$). Similarly, the Human transcript performed significantly better than the Random baseline as well (using a paired t-test $p<0.01$). Furthermore, the rates of Human and RMVS were similar with no statistically significant difference. To conclude, the results support the experiment's hypothesis. People were not able to distinguish between the Human suspect and the RMVS. However, they were able to easily distinguish between the Random baseline and the RMVS as well as between the Random baseline and the Human.

\section{Conclusion}

In this paper the Virtual-Suspect system was presented, which consists of a configurable simulation and the Virtual-Suspect Response Model (RMVS). The \textit{Scenario Configuration} and \textit{Interrogative Interviewing} databases enable the simultaneous simulation of multiple interrogation cases. Furthermore, the Virtual-Suspect's personality profile can be configured which is based on Eysenck's PEN Model. In addition, the Virtual-Suspect Response Model (RMVS) was presented. For every interrogator's statement, the RMVS updates the internal-state vector, converts it into the Mental-Integrity vector which determines the probability
of responding with a truthful, deceptive or a neutral response. Lastly, a perceptual experiment was conducted comparing the RMVS, Human and Random responses. The results suggest that humans have difficulty differentiating between simulations generated by the RMVS and those of a human.

\bibliography{iva}
\end{document}